\documentclass[runningheads]{llncs}
\usepackage[T1]{fontenc}
\usepackage{graphicx}
\usepackage{xcolor}

\begin{document}
\title{Educational robotics beyond ROS: a custom library for direct transfer of Webots controllers in the Turtlebot3}
%
%
\author{Jaime P. Pérez-Moro\inst{1}\orcidID{0009-0002-3143-2386} \and
Alejandro Romero\inst{1}\orcidID{0000-0002-0507-8320} \and
Hector Quintian\inst{2}\orcidID{0000-0002-0268-7999} \and 
Francisco Bellas\inst{1}\orcidID{0000-0001-6043-1468}}
\authorrunning{J. P. Pérez et al.}
\titlerunning{Direct transfer of Webots controllers in the Turtlebot3}

\institute{Integrated Group for Engineering Research (GII), CITIC research center, Universidade da Coruña, A Coruña, Spain\\
\email{\{j.p.perez,alejandro.romero.montero, francisco.bellas\}@udc.es}\and
Grupo de investigación Ciencia y Técnica Cibernética (CTC), CITIC research center, Universidade da Coruña, A Coruña, Spain\\
\email{hector.quintian@udc.es}}

\maketitle              
\begin{abstract}
Robot Operating System (ROS) is becoming a standard in robotics courses at higher education, but its installation, configuration and distributed programming model can introduce a significant entry barrier in the new landscape of AI-based master and degree studies, where many students do not have a strong informatics background. This paper presents an educational experience that enables students to program a TurtleBot3 Burger directly in Python, while preserving a direct transfer path from simulation to the physical robot. The approach is based on a custom library that mirrors the Webots API, allowing the same controller code to be executed in the simulator and on the real TurtleBot3 with minimal modification. The library provides access to the robot motors, wheel encoders, inertial sensors and LiDAR, and is complemented by scripts that automate deployment and execution on the physical platform. It was used in a master-level course on introductory autonomous vehicles, and initial student feedback suggests that the approach reduces the perceived complexity of ROS-related infrastructure and helps learners focus on algorithmic logic and real-world behaviour.

\keywords{Educational robotics  \and Simulation-to-reality transfer \and ROS}
\end{abstract}
\section{Introduction}

The use of ROS \cite{ros} in technical subjects and courses related with autonomous robotics at higher education is becoming a standard \cite{ryalat}\cite{Mamatnabiyev}. Its open-access nature, hardware abstraction and huge community of contributors make ROS as the optimal choice for teachers that aim to provide students with the most solid and long-term training in robotics. In addition, the current offer of affordable and versatile real platforms for educational/research purposes is mainly ROS-compatible and even ROS-exclusive, like those from Unitree, Universal Robots, PAL or Clearpath Robotics. 

Since the ``explosion'' of generative AI in 2022, the interest of students in titles related with Artificial Intelligence (AI) has notably increased and, consequently, the offer of new titles related with this area has raised \cite{Crompton}. Many of these new students and titles do not have or require the high technical background associated with ROS, although they need learning about autonomous robotics in a solid and sound way. There are notably relevant initiatives that aim to ``smooth'' the ROS learning curve, such as the web-based Unibotics framework \cite{unibotics}\cite{roldan}, but the real situation in many of these new courses is that installing Linux or relying on virtual machines is complicated, mainly due to time constraints in the subjects, and the lack of support from the institutions.

With this background, the current paper describes a custom library developed at the University of Coruña (UDC), in Spain, together with the educational experience in which it was applied. The library was created for a technical master program named ``Industrial Informatics and Robotics'' \cite{UDC}. This subject is a clear example of the situation presented above, as it is mainly technical, but the number of students attending it with limited programming and computer science background increases every year. In particular, the experience described here was carried out in the realm of a subject named ``Introductory Autonomous Vehicles'', where students learn the fundamentals of this field, including topics like perception, mapping, localization and control.

The practical part of the subject implies programming the response of an autonomous robot, in simulation and in the real-world, requiring a minimum proficiency level as a computer scientist. The specific platform existing in the UDC for this subject is the TurtleBot3 Burger, a ROS-exclusive platform that is shared with other subjects in the same master. To provide this heterogeneous group of students with a sound and simple educational environment that allow them to practice the subject fundamentals without relying on ROS, we decided to develop a custom library so they can program it in a simpler way using the Webots simulation environment \cite{Webots}. 

This work describes the implementation details of this library, how it was validated in the simulator and the real robot, and its application during academic year 2025/26 within the aforementioned subject.

\section{Custom library development}
This section describes the development of the custom library. First, the Webots simulation environment is presented, justifying its selection for this work. Next, the design and implementation of the library are detailed, covering the low-level access to the TurtleBot3 Burger hardware and the way its interface mirrors the Webots API so that the same controller runs unchanged in simulation and on the real robot. Finally, the operational validation carried out before using the library in a teaching setting is reported.

\subsection{Webots simulation environment}
Webots~\cite{Webots} is an open-source and widely adopted simulation platform that was selected for this course after evaluating several of the alternatives commonly used in educational robotics~\cite{Tselegkaridis}, based on a combination of features well-suited to an educational context with a heterogeneous student background. Its physics engine, the Open Dynamics Engine (ODE), enables realistic simulations supporting collisions, friction and gravity, which are essential for an accurate reproduction of robot behaviour. It is also flexible in terms of programming languages, since controllers can be programmed in Python, C, C++, Java or MATLAB, a key advantage for groups with limited programming background.

Webots further supports many popular robot models out of the box, including the TurtleBot3 Burger, which avoids building models from scratch and makes experimentation with the target platform straightforward. World creation, controller programming and debugging are all carried out from a single unified interface, and its extensive documentation and active community lower the entry barrier for new users. The accuracy of motion simulation and the comparison with other simulators that supported this choice are discussed in \cite{Farley}.

\subsection{Design and implementation}
\begin{figure}
\centering
\includegraphics[width=12cm]{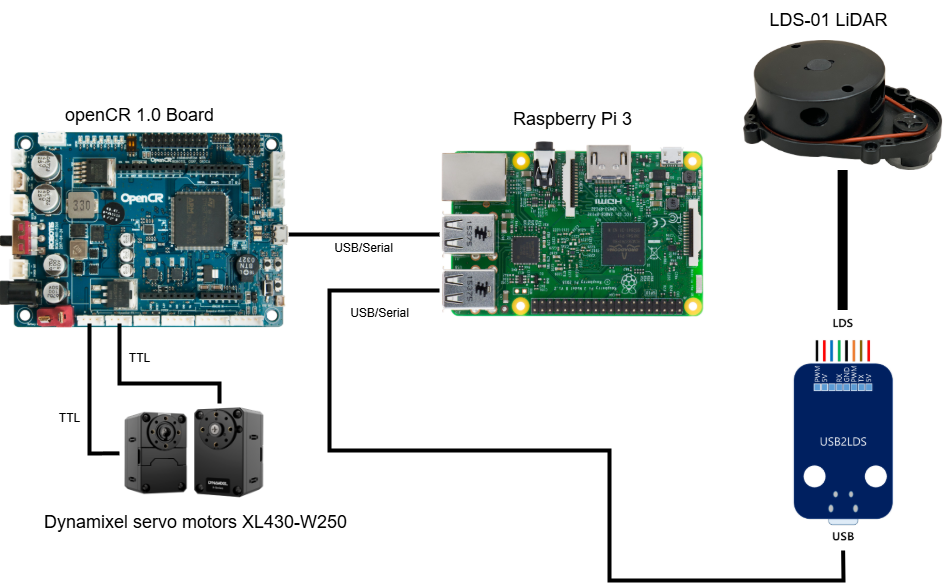}
\caption{TB3 Burger connection diagram.} \label{conexion_tb3}
\end{figure}

The development started by addressing low-level access to the TurtleBot3 Burger, controlling its components directly from Python and without relying on ROS. The robot integrates a Raspberry Pi 3 as the main computer, two Dynamixel XL430-W250 servomotors driving the wheels, an OpenCR 1.0 board hosting an inertial measurement unit (IMU), and an LDS-01 LiDAR, Fig. \ref{conexion_tb3} illustrates how all of these components are connected together.

The servomotors connect to the Raspberry Pi 3 through the OpenCR 1.0 board over a serial port. As the board ships with a ROS-oriented firmware, it was reprogrammed using the Arduino IDE with custom code based on the official Dynamixel2Arduino library~\cite{Dynamixel2Arduino}. This firmware runs a main loop that monitors the serial input for commands from the Raspberry Pi 3 and reacts to two types of request: setting the velocity of a wheel, and reading a wheel encoder value, which is returned over the serial link. 

Communication on the Raspberry Pi side was handled with the \texttt{pyserial} library. Access to the IMU followed a similar approach, extending the firmware with reading functions adapted from the official OpenCR examples to expose raw accelerometer and gyroscope data and the computed roll, pitch and yaw values. The LiDAR LDS-01, by contrast, is connected through an independent USB2LDS board that did not require reprogramming, and its data are read directly from the assigned serial port by a dedicated class that performs the acquisition in a separate thread.

These low-level components were then unified into a single library whose key design decision was to mirror the Webots API~\cite{WebotsAPI} as closely as possible, so that the same controller code could run unchanged both in simulation and real robot. To this end, the library functions were given the same names, behaviours and return values as their Webots counterparts, following a modular, object-oriented architecture in which each component is encapsulated in its own class. 

\begin{figure}
\centering
\includegraphics[width=11cm]{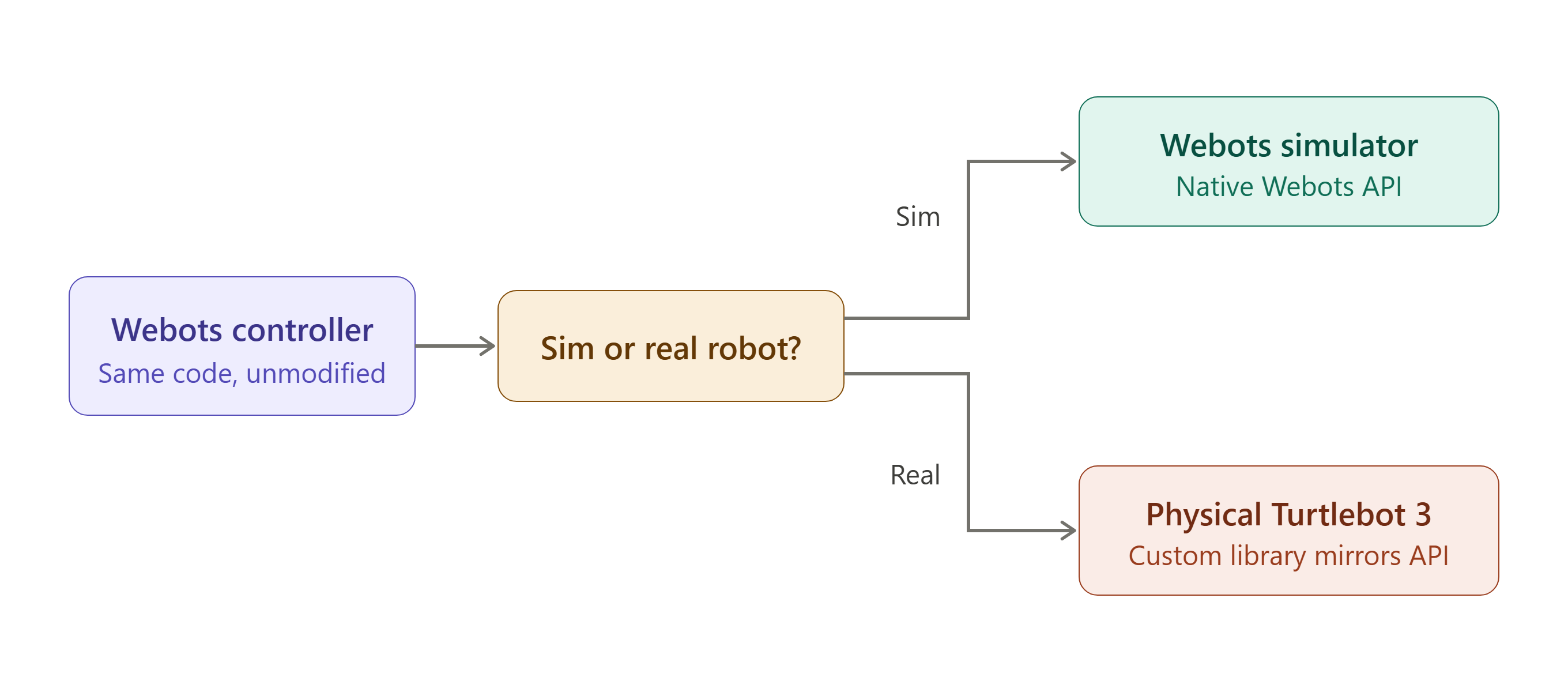}
\caption{Execution flow of a Webots controller.} \label{diagram_fig}
\end{figure}

A \texttt{Robot} class acts as the main entry point, centralising access to the devices and exposing a \texttt{step(timestep)} method that emulates the Webots simulation timestep so that sensor readings and command execution occur at consistent time intervals on the physical robot. The remaining classes (\texttt{Motor}, \texttt{PositionSensor}, \texttt{Accelerometer}, \texttt{Gyro}, \texttt{InertialUnit} and \texttt{Lidar}) follow the same principle and replicate the behaviour of their Webots equivalents, for instance converting raw encoder counts into radians as Webots does. In practice, a small code block at the beginning of any Webots controller checks whether the Webots \texttt{Robot} class is available; if it is not, indicating execution on the physical hardware, the custom library is imported instead, so no further modification of the controller is needed as seen in Fig. \ref{diagram_fig}.

Finally, the transfer and execution of code on the real robot were automated through a set of scripts working on both Linux and Windows, removing the need to manually establish an SSH connection and copy files. The scripts rely on \texttt{scp} and \texttt{ssh}, and offer three commands: \emph{deploy}, which transfers the Python controller to the TurtleBot3 Burger given its IP address; \emph{run}, which transfers and executes the code on the robot; and \emph{clean}, which removes the transferred file. The library, scripts and documentation are openly available in our GitHub repository.\footnote{TurtleBot3-NoROS-Library: \url{https://github.com/jpperezm/TurtleBot3-NoROS-Library}.}

For reproducibility, the software stack used in the reported experience is the following: Webots R2025a on the students' computers, running either Windows or Linux; Ubuntu Server 22.04 LTS on the Raspberry Pi 3 of the TurtleBot3, with Python 3.10 and \texttt{pyserial} 3.5; and Arduino IDE 2.3 together with Dynamixel2Arduino 0.8.1 for building the OpenCR 1.0 firmware.

\subsection{Operational validation}
Before relying on the library in a teaching setting, each component was checked individually, without a control loop, to confirm that the same controller produced equivalent results in simulation and on the physical robot. Three checks were performed: moving the robot 25~cm forward using the wheel encoders, commanding a 90-degree left turn governed by the gyroscope, and advancing towards an obstacle until a target LiDAR reading was reached. In every case the simulated and real runs were close and, importantly, highly repeatable, which is what matters for transferring a controller between the two. The aim was not to characterise sensor error precisely, but to confirm that each component behaves consistently enough for code developed in simulation to carry over to the real robot.

All these tests were confirmed by a final experiment that combined all the components in a single task: the robot had to move 25~cm forward using the encoders, turn 45 degrees to the left using the gyroscope, and then advance while avoiding obstacles with the LiDAR, reacting with a 90-degree turn when an obstacle was detected within 20~cm. The controller was first developed and tuned in Webots, the laboratory was arranged to match the simulated scenario, and the same code was then transferred to the physical TurtleBot3 using the automated scripts without any modification. The robot reproduced the programmed sequence reliably on both platforms\footnote{A video comparing the execution of the same controller in simulation and on the real robot is available at \url{https://youtu.be/bzEeayc3Oj4}.}, confirming that controllers written for Webots can be transferred directly to the real robot and that the Webots--TurtleBot3 combination is a viable solution for teaching robotics without ROS.

\begin{figure}
\centering
\includegraphics[width=11cm]{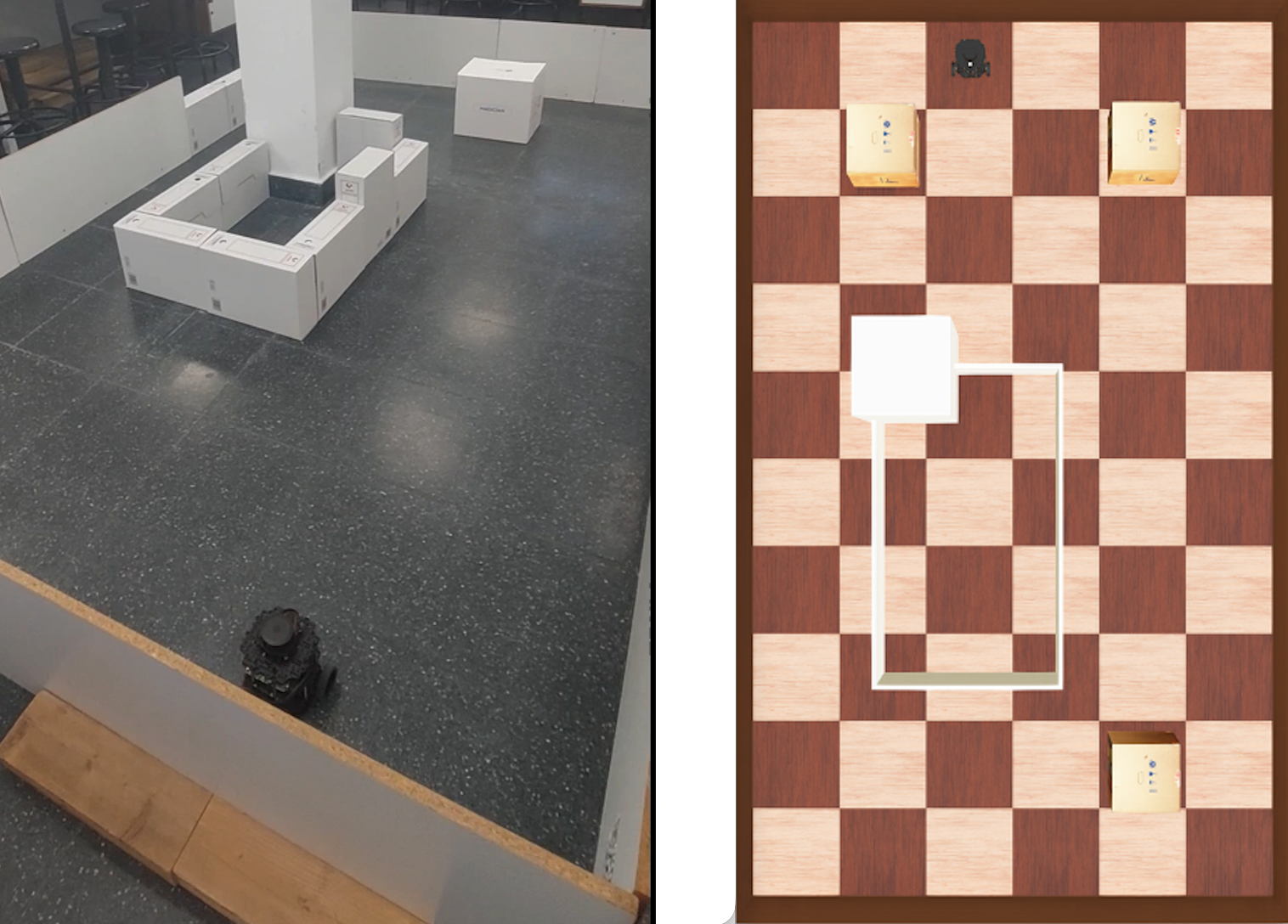}
\caption{Real and simulated environment used for the activity.} \label{environment_fig}
\end{figure}

\section{Results}

This section describes the educational experience that was carried out in the scope of the ``Introductory Autonomous Vehicles'' subject from the ``Industrial Informatics and Robotics'' master during course 2025/26, where the custom library was used. Six students were enrolled in the course this year, as it is not an elective subject, so the master's students are distributed across different courses. The practical task they had to solve was the following:
\begin{itemize}
    \item Design of a deliberative autonomous architecture, so the TurtleBot3 Burger robot can navigate from one point to another specified by the user, avoiding obstacles and following the shortest path. To this end, the architecture must solve 3 independent problems: (1) map creation (2) path planning (3) navigation. It must be solved in Webots and also in the real robot.
    \item Teachers provide 3 different environments in simulation, and setup one of them in the real world. They are created as grids, with known maximum dimensions, $6 \times 10$ cells, with each cell measuring $40 \times 40$ cm. The robot starts from a base cell located next to a wall, but its absolute position and initial orientation are not known in advance, Fig. \ref{environment_fig} shows one of the simulated environments, and the real one.
    \item Although the position of the base cell within the environment is not given to the robot, the initial pose is not estimated globally either. The robot is always placed in the base cell in the same way as in the simulated world, and it takes its own initial heading as the reference direction of the map it builds, so that the map frame is defined relative to the starting pose. Students therefore do not have to solve a global initial-localisation problem: they only have to track position and orientation by odometry, which should be identical in simulation and on the real robot.
    \item Localization must be carried out in two ways. First, using the Webots global localisation through the ``supervisor'' feature, to simplify development and debugging. The final implementation must replace this with local odometry based on the robot’s wheel encoders and gyroscope, so that the same solution can later be transferred to the physical robot.
    \item During the map creation phase, the robot builds a binary occupancy grid, distinguishing between free and occupied cells. LiDAR readings are taken with the robot stopped at the centre of each cell, allowing the system to detect walls or obstacles in adjacent cells and update the internal map.
    \item Once the map has been created, the robot must use it to compute an optimal path from its current position to a user-defined destination cell. This second stage assumes a static environment, so the occupancy grid generated during exploration remains valid for navigation.
    \item Before using the real robot, students test the complete behaviour in Webots: discrete motion, map updating, localisation, obstacle detection and path following. The goal is to ensure that the algorithm works reliably in the simulated environment shown in the accompanying figure.
    \item After validation in simulation, the students transfer their developed code to the TurtleBot3. This step requires adapting the implementation so that it uses the real robot interfaces instead of the simulated Webots devices.
\end{itemize}

\begin{table}[ht]
\centering
\begin{tabular}{p{\textwidth}}
\hline
\textbf{Based on your knowledge and experience about ROS, what main advantages did you find in programming and running the TurtleBot3 directly in pure Python?} \\
\hline
- Simpler and more focused on the control logic. \\
\hline
- When programming directly in Python, you do not need to think about the distribution of nodes or communication mechanisms, so the programming style is closer to the type of programming we use in the rest of the subjects. \\
\hline
- I do not have to think about packages, services, messages, publishing, subscription, etc. \\
\hline
- The main advantage I found in Python is the ease of integrating code libraries, such as NumPy or priority queues. \\
\hline
- The simplicity of using only a Python script and not having to deal with ROS packages. \\
\hline
\end{tabular}

\vspace{0.3cm}

\caption{Students' responses on the advantages of programming and running the TurtleBot3 directly in Python. It must pointed out that this group of students has basic ROS knowledge due to a theoretical class they receive in other subject of the master.}
\label{tab:python_turtlebot3_advantages}
\end{table}

At the end of the semester, students were asked to fill in a survey to get their feedback regarding library usage. Table \ref{tab:python_turtlebot3_advantages} contains qualitative opinions with regards to the advantages of using this library as compared to ROS. It can be observed that most of the students confirm the simplicity of this approach as the main advantage.

\begin{figure}
\includegraphics[width=\textwidth]{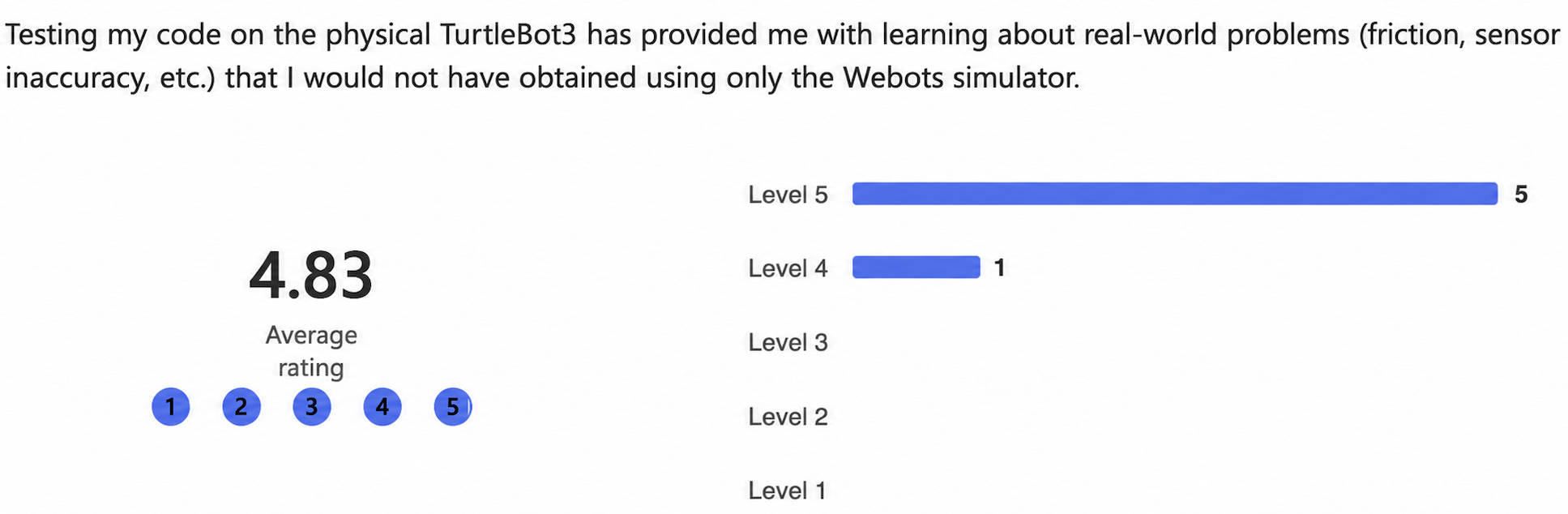}
\caption{Student's answer related with the use of the real robot.} \label{survey_fig}
\end{figure}

Although it was not the main objective here, Fig.\ref{survey_fig} shows an interesting result we have obtained from this experience. Even in the case that these students did not have a deep informatics background, they highlight the use of the real robot as compared to only the simulated one. We want to point out this, because many new courses at this level tend to avoid physical robots due to its complexity, but students' opinion is really positive towards its use.

\section{Conclusions}

This paper presented a custom Python library that enables the direct transfer of Webots controllers to a physical TurtleBot3 Burger without relying on ROS. The objective is not to replace ROS as a standard robotics middleware, but to provide a simpler educational pathway for courses where the main focus is autonomous robotics rather than middleware configuration. The library was applied in a master-level course on ``Introductory Autonomous Vehicles'', where students addressed a complete task involving map creation, path planning and navigation in both simulated and real environments. Student feedback suggests that the approach reduces the perceived complexity associated with ROS packages, messages, publishers, subscribers and distributed nodes, allowing learners to focus more directly on control and navigation logic.

Regarding scalability, the library currently covers the devices required by
the course: wheel motors and encoders, IMU and LiDAR. The object-oriented design
encapsulates each device in its own class, so adding a new one connected through
the serial or USB interfaces only requires a class mirroring the corresponding
Webots node. Higher-bandwidth sensors such as the RGB-D cameras of other
TurtleBot3 variants would need a different acquisition path and are left as
future work.

\begin{credits}
\subsubsection{\ackname} This work was supported by the Spanish Ministry of Science, Innovation and Universities through the grant FPI-UDC2025/06 funded by PID2024-162223OB-I00, and CITIC (Xunta de Galicia - Convenio para o desenvolvemento de accións estratéxicas de I+D+i 2025-2026).

\subsubsection{\discintname}
The authors have no competing interests to declare that are relevant to the content of this article. 
\end{credits}

%
%
%

\begin{thebibliography}{8}
\bibitem{ros}
Quigley, M., Conley, K., Gerkey, B., Faust, J., Foote, T., Leibs, J., Wheeler, R., Ng, A.Y.: ROS: an open-source Robot Operating System. In: ICRA Workshop on Open Source Software. vol. 3, p. 5 (2009)

\bibitem{ryalat}
Ryalat, M., Almtireen, N., Al-refai, G., Elmoaqet, H., Rawashdeh, N.: Research and Education in Robotics: A Comprehensive Review, Trends, Challenges, and Future Directions. Journal of Sensor and Actuator Networks, 14, 76. (2025) \doi{10.3390/jsan14040076} 

\bibitem{Mamatnabiyev}
Mamatnabiyev, Z., Chronis, C., Varlamis, I., Himeur, Y., Zhaparov, M.: A Holistic Approach to Use Educational Robots for Supporting Computer Science Courses. Computers, 13, 102 (2024) \doi{10.3390/computers13040102}

\bibitem{Crompton}
Crompton, H., Burke, D. Artificial intelligence in higher education: the state of the field. Int J Educ Technol High Educ 20, 22 (2023). \doi{10.1186/s41239-023-00392-8}

\bibitem{unibotics}
Roldán-Álvarez, D., Cañas, J. M., Valladares, D., Arias-Perez, P., Mahna, S.: Unibotics: open ROS-based online framework for practical learning of robotics in higher education. Multimedia Tools and Applications, 83(17), 52841-52866 (2024)

\bibitem{roldan}
Chen, L. et al. (2025). Teaching Service Robotics with ROS and Unibotics Web Framework in Higher Education. In: Balogh, R., Obdržálek, D., Fachantidis, N. (eds) Robotics in Education. RiE 2025. Lecture Notes in Networks and Systems, vol 1544. Springer, Cham. \doi{10.1007/978-3-031-98762-5_43}

\bibitem{UDC}
University of Coruña, Industrial Informatics and Robotics Master, accessed 10/06/2025, 
\url{https://estudos.udc.es/en/study/detail/4556V01}

\bibitem{Webots}
Cyberbotics Ltd.: Webots: open-source robot simulator. \url{https://cyberbotics.com}, last accessed 10/06/2025

\bibitem{Tselegkaridis}
Tselegkaridis, S., Sapounidis, T.: Simulators in Educational Robotics: A Review. Education Sciences 11, 11 (2021) \doi{10.3390/educsci11010011}

\bibitem{Farley}
Farley, A., Wang, J., Marshall, J. A.: How to pick a mobile robot simulator: A quantitative comparison of CoppeliaSim, Gazebo, MORSE and Webots with a focus on accuracy of motion. Simulation Modelling Practice and Theory, 120, (2022). 102629.

\bibitem{Dynamixel2Arduino}
ROBOTIS: Dynamixel2Arduino: DYNAMIXEL protocol library for Arduino. \url{https://github.com/ROBOTIS-GIT/Dynamixel2Arduino}, last accessed 10/06/2025

\bibitem{WebotsAPI}
Cyberbotics Ltd.: Webots Reference Manual --- API. \url{https://cyberbotics.com/doc/reference/nodes-and-api-functions}, last accessed 10/06/2025
\end{thebibliography}

\end{document}